# Seeing poverty from space, how much can it be tuned?

Tomas Sako[1], Arturo Jr M. Martinez[2]

[1] Corresponding author. Email: tomas.sako.mgr@gmail.com
[2] Asian Development Bank

## Abstract

Since the United Nations launched the Sustainable Development Goals (SDG) in 2015, numerous universities, NGOs and other organizations have attempted to develop tools for monitoring worldwide progress in achieving them. Led by advancements in the fields of earth observation techniques, data sciences and the emergence of artificial intelligence, a number of research teams have developed innovative tools for highlighting areas of vulnerability and tracking the implementation of SDG targets.
In this paper we demonstrate that individuals with no organizational affiliation and equipped only with common hardware, publicly available datasets and cloud-based computing services can participate in the improvement of predicting machine-learning-based approaches to predicting local poverty levels in a given agro-ecological environment.

The approach builds upon several pioneering efforts over the last five years related to mapping poverty by deep learning to process satellite imagery and "ground-truth" data from the field to link features with incidence of poverty in a particular context. The approach employs new methods for object identification in order to optimize the modeled results and achieve significantly high accuracy. A key goal of the project was to intentionally keep costs as low as possible - by using freely available resources - so that citizen scientists, students and organizations could replicate the method in other areas of interest. Moreover, for simplicity, the input data used were derived from just a handful of sources (involving only earth observation and population headcounts). The results of the project could therefore certainly be strengthened further through the integration of proprietary data from social networks, mobile phone providers, and other sources.

## Related work

Several methods for estimating and predicting poverty at a "granular" level have been pioneered and further refined in recent years. In general, the original method [1] involved training a Convolutional Neural Network (CNN) to predict nightlight intensity of a given geographic area based on the daytime images of the same area. Based on the well-tested assumption that nightlight levels serve as a relatively good proxy for economic livelihood, the initial approach was further adapted for predicting poverty by applying the transfer learning method.
The field was further advanced when researchers in [2] proposed applying an object detection model on very high resolution daytime images. Once the model could detect absolute counts of real objects like cars, planes, trucks or maritime vessels in a given satellite image - this led to the creation of new training data sets for poverty modeling. The method with object detection worked sufficiently well, with additional benefit for policymakers of being easier to interpret the results of predictions.

[3] trained a poverty prediction model using not only satellite images, but also data sets from different domains such as social media, mobile phone data and other sources.

The increasingly robust results from the evolutions of the constellation of methods briefly discussed above suggest that curating and combining of human-generated data sets from various domains coupled with the processing and classification prowess of advanced computer vision algorithms defines the current state-of-the-art in the field of wealth modeling, offering the way how to maximize the predictive performance of poverty models.

## Input data

For daytime imagery we used Sentinel 2 [11] low resolution imagery with full coverage of the country area. Specifically, this involved 13 000 images consisting of 384 pixels each, with a spatial resolution 10 meters per pixel. In addition, the team supplemented Sentinel data with Google Earth images [6] which provided a considerably closer view of the target areas, with spatial resolution 0.3 metres per pixel (resp. 0.9m/px after triple downsampling). Nevertheless, as was experienced by other research teams [12], due to limitations and restrictions on Google Earth data, we used these data only to perform the subset of experiments on a small sample of the Philippines area (roughly 2% of the land). Due to mass download restrictions on the Google Earth platform, only 3740 images with high pixel resolution were downloaded and further processing tasks - including splitting the images into tens of thousands smaller training image tiles (training chips) - were performed locally.

## Data processing

In Google Earth we defined areas of interest (AOIs) and downloaded very high resolution (VHR) images Fig[2].

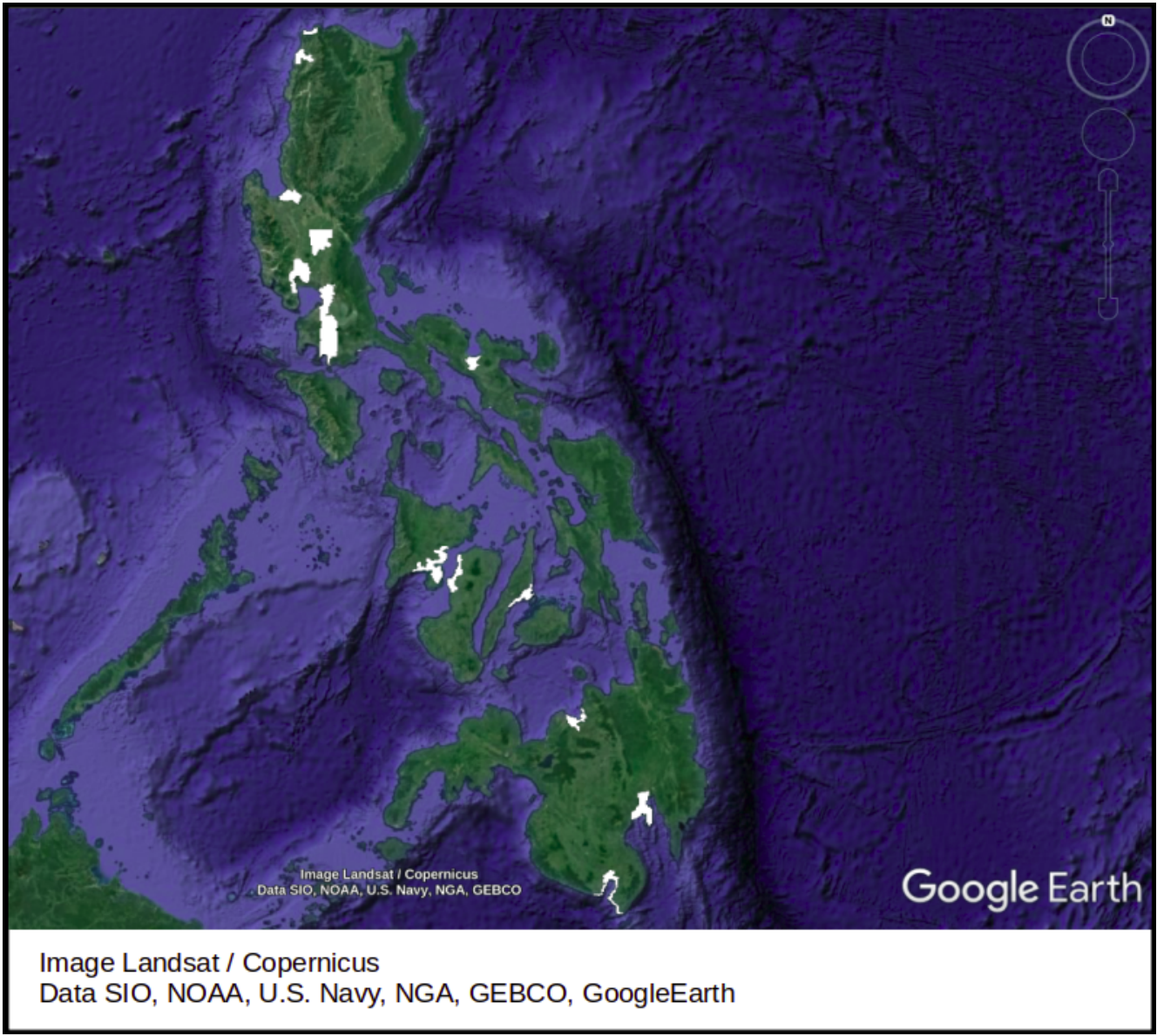

Fig[2] Areas of interest in white.

## Training - Validation - Test splits

The results of the linear poverty models developed by this project were calculated based on poverty prediction on roughly 20% of all selected provincies. This holdout test set was chosen randomly and is generally consistent, in other words, the same test set was used throughout all of the experiments. Moreover, as noted above, very high resolution (VHR) satellite imagery was collected for a subset of target areas (180 provinces were collected in total, out of which 43 provinces were assigned to the test set.
Validation sets (needed for the monitoring of the training process) were selected by each individual framework on its own.

On the other hand, the object detector's performance is evaluated by mAP.5 and mAP.5:.95 metric functions. While xView daytime image data set covers several areas all around the world, the same split as for classification and regression models could not have been applied. Before the object detection training started, we had randomly created holdout test set and checked that there was a sufficient amount of object instances for each object prediction class. As for training-validation split, we just took the default split in the original xView folder structure.

Once trained, the detector ran inference on non-annotated VHR images of Philippines, individual counts were imported by extract-transform-load (ETL) feed to the database and aggregated for each of 180 studied provinces. This data set was used for the linear poverty modeling step, while preserving the previously mentioned province split (43 test provinces out of all 180).

## Centroid grid

We extracted GPS locations from the center of each pixel in VIIRS nighttime light (VNL) data set and took only a subset of them located in manually defined "areas of interest" (AOI), see Fig[2]. The area on the ground covered by every training image was estimated as the average of the VNL pixel size at the northernmost (430 meters) and southernmost (470 meters) part of the Philippines which resulted in 450m x 450m area covering images. The official image resolution of VNL is roughly 500m at the Equator [14]. See in Fig [3] the alignment of our training images with original VNL pixels.

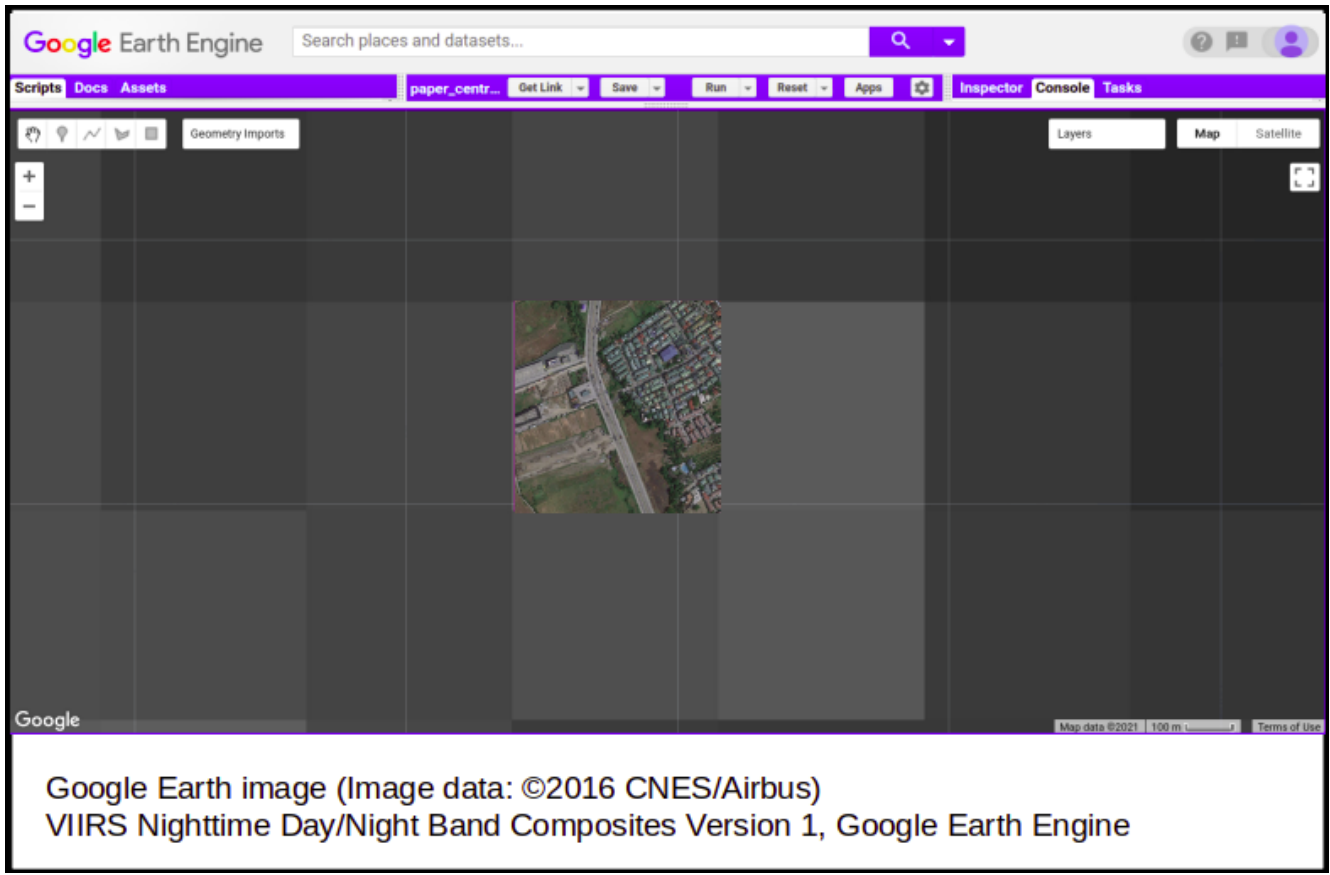

Fig[3] original VNL pixels vs training image sample

Locations of images for download were filtered based on following criteria:

1. Worldpop data [8] showing at least 2 people living in the area covered by image in 2015
2. Google Earth has available clear VHR image of this area from period 2015-2016

To optimize numerical errors by minimizing the count of aggregations, rounding and other numerical operations, we attempted to align a centroid list to the original VNL's grid of pixels in TIF format.

# Modeling

# Training of CNNs

All classification and regression convolution neural networks (CNN) were trained in Fastai[7] based on the approach with nightlights described in [1]. Fastai framework has been widely adopted by the research community because a full machine learning pipeline including training, validating and testing can be designed efficiently and consistently.

Classification models for predicting nightlight intensity classes were calculated by summation of nightlight intensity values within the geographical area covered by each daytime satellite image. Once having aggregated nightlight value, the Gaussian mixture model was applied to cluster all images into three nightlight classes, which then, in turn, became target prediction classes during the classification training process.

For nightlight data, the VIIRS sensor[4], [5] was selected as it currently provides the highest spatial resolution and offers a number of improvements over Operational Line-Scan System [10]. Annual composites for the year 2015 were cropped only for the area of the Philippines. We picked version 1 and version 2 data sets during our experiments.

At the same time, and in addition to classes, regression models were also trained to predict the continuous variable – real nightlight intensities, hence different loss and metric functions had to be set up.

## Visualization of the features from the classification model

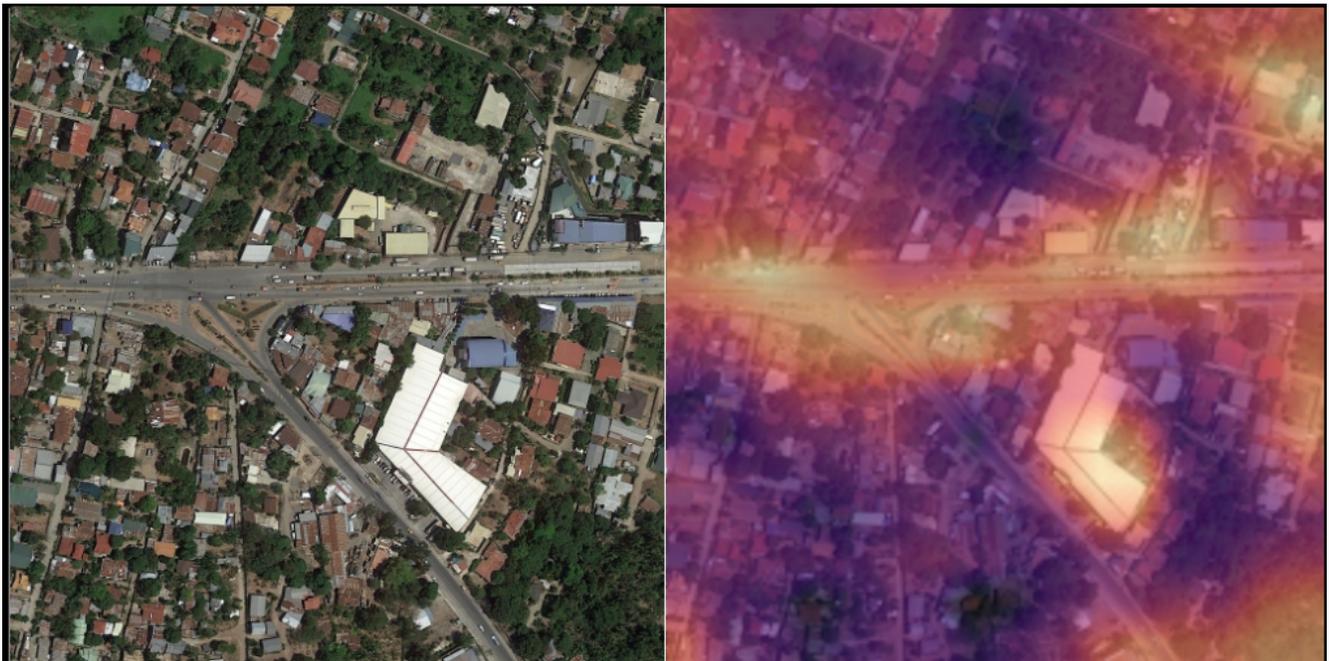

Google Earth image (Image data: ©2016 CNES/Airbus)  vs  Fastai heatmap

Fig[4] original daytime satellite images from Google Earth on the left and overlay of activation maps onto original images on the right

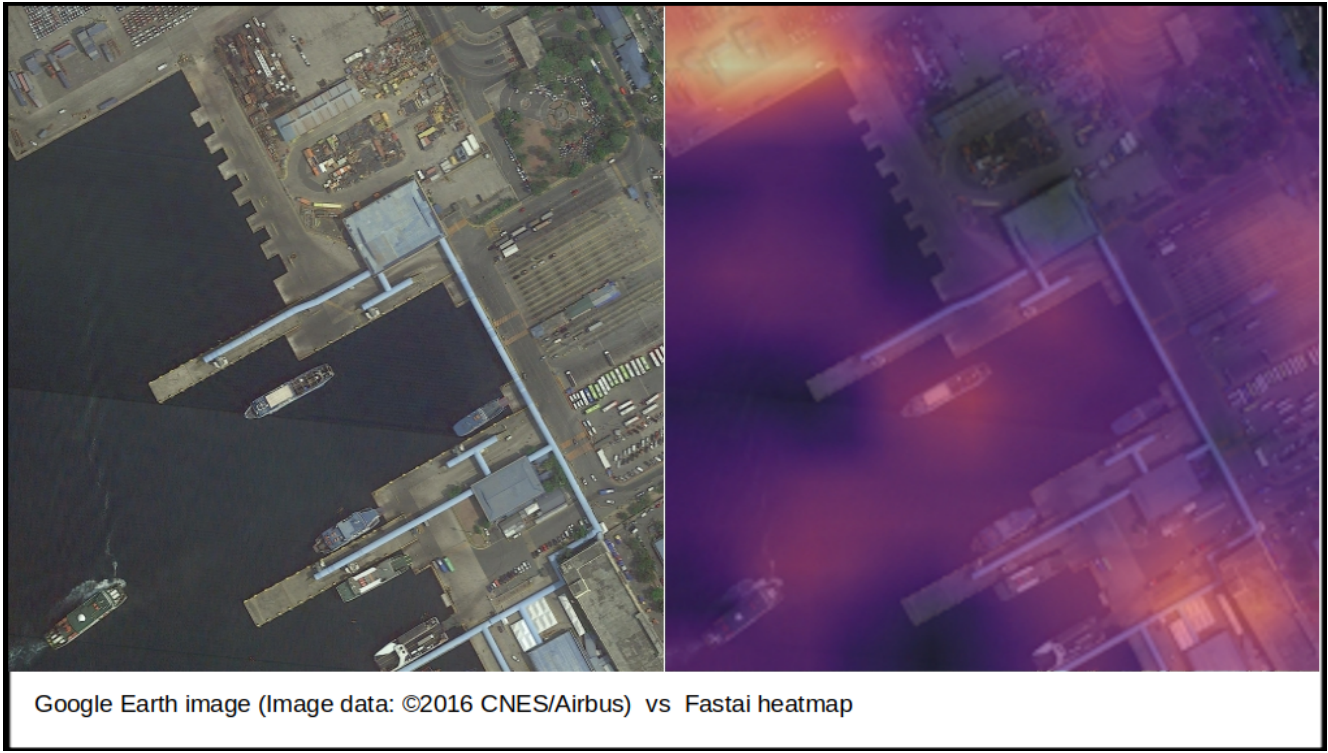

Fig[5] each heatmap highlights the parts of the image with last activations - meaning features multiplied by last weight matrix - the strongest activations are shown in white

This analysis of heatmaps covers the prediction of nighlight classes - not poverty - so if there is stronger emphasis on bigger buildings, it could mean that model learnt that big buildings simply "produce" more high intensive light at night, the same with water surfaces and other objects.

We observed that CNN learnt that certain man-made structures strongly impact nightlight intensity in their respective area similarly as in [1]. Our analysis included high resolution imagery and revealed that even individual large objects as ships are considered by CNN as significant features.

Besides these, we detected big buildings and their vehicle lots, engineering and construction sites, highways and container warehouses as another important features for CNN's prediction of high nightlight intensity.

# Training of the object detector

We also trained the object detection model based on the approach described in [2]. We performed training on a public xView annotated data set where we grouped individual object classes to 10 parent classes, see Tab[1].

| Fixed-Wing Aircraft | Passenger Vehicle | Truck | Railway Vehicle | Maritime Vessel | Engineering Vehicle | Building | Helipad | Vehicle Lot | Construction Site | None |
|---|---|---|---|---|---|---|---|---|---|---|
| Small Aircraft | Small Car | Pickup Truck | Passenger Car | Motoboat | Tower Crane | Hut/Tent | | | | Pylon |
| Cargo Plane | Bus | Utility Truck | Cargo Car | Sailboat | Container Crane | Shed | | | | Shipping Container |
| | | Cargo Truck | Flat Car | Tugboat | Reach Stacker | Aircraft Hangar | | | | Shipping Container Lot |
| | | Truck w/Box | Tank Car | Barge | Straddle Carrier | Damaged Building | | | | Storage Tank |
| | | Truck Tractor | Locomotive | Fishing Vessel | Mobile Crane | Facility | | | | Tower Structure |
| | | Trailer | | Ferry | Dump Truck | | | | | Helicopter |
| | | Truck w/Flatbed | | Yacht | Haul Truck | | | | | |
| | | Truck w/Liquid | | Container Ship | Scraper/Tractor | | | | | |
| | | | | Oil Tanker | Front Loader | | | | | |
| | | | | | Excavator | | | | | |
| | | | | | Cement Mixer | | | | | |
| | | | | | Ground Grader | | | | | |
| | | | | | Crane Truck | | | | | |

Tab[1] xView parent object classes

Publicly available implementation of Yolo5 object detector from Ultralytics[13] was applied during all modeling phases.

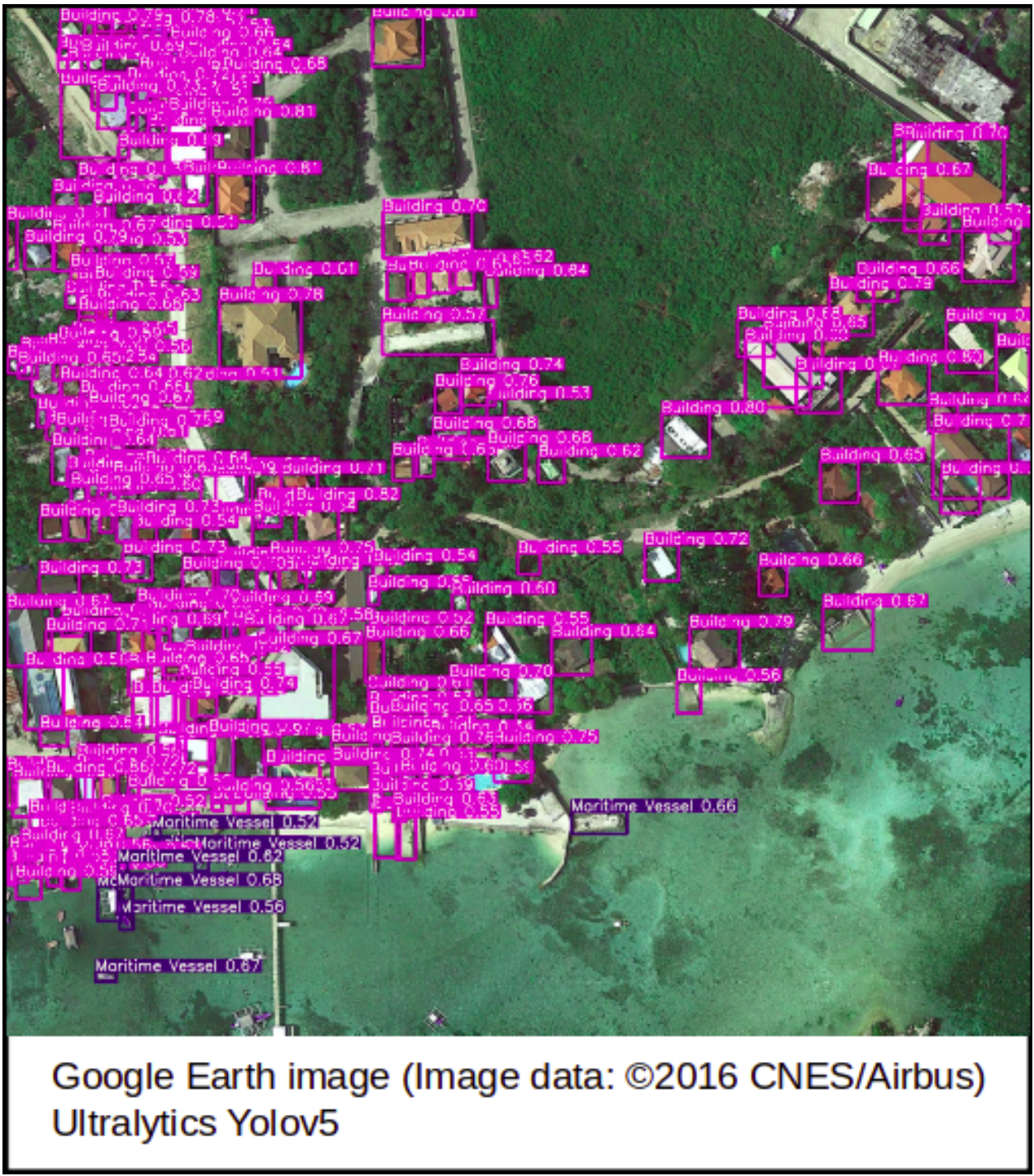

Fig[6] Example annotations detected by our model.

## Processing of annotations

xView annotations were checked for correctness with the following criteria:

1. images with single incorrect annotation were preserved in data set, incorrect annotation was removed
2. images with multiple incorrect annotations were completely removed

As the next processing step, the format of annotations was converted to Yolo format.

## Dealing with imbalanced object type occurrences

The occurrences of various object types in xView data set are highly imbalanced. To resolve this, we assigned individual weight to each class according to its cardinality within the whole xView data set, where the rarest classes received the highest weights. Then we split each original xView image into 16 equal size quadrants, see Fig[7]. Based on summed object's weights within quadrants we assigned probabilities to each quadrant, see Tab[2]. Hence, instead of random sampling, we sampled small training image tiles (training chips) from a specific quadrant based on calculated probabilities. This approach increased the occurrences of rare object types in our training set and therefore helped to mitigate class imbalancement.

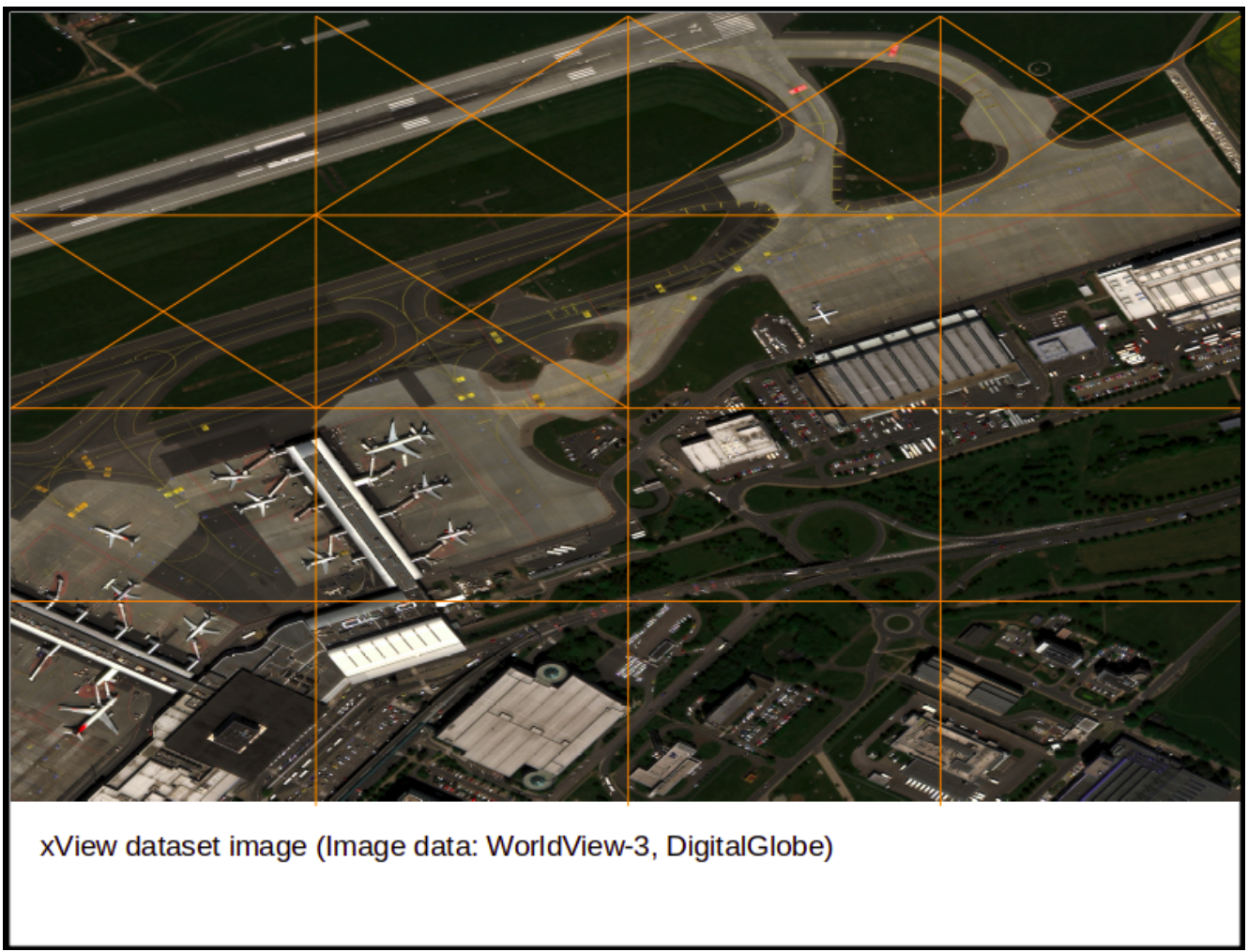

Fig[7] Five highlighted quadrants contain no object annotations, so there is zero probability of sampling small chips for training. Even if there are only a few objects in a given quadrant, it can gain similarly high sum of weights as quadrant with i.e. lots of passenger cars, which is caused by different weights for various object types.

| | orig_filename | row_i | col_j | topleft_x | topleft_y | orig_width | orig_height | quad_width | quad_height | sum_w | prob | prob_from | prob_to |
|---|---|---|---|---|---|---|---|---|---|---|---|---|---|
| 1 | 311.jpg | 0 | 0 | 0 | 0 | 4763 | 3064 | 1190 | 766 | 15 | 0.00070788 | 0 | 0.00070788 |
| 2 | 311.jpg | 0 | 1 | 0 | 766 | 4763 | 3064 | 1190 | 766 | 0 | 0 | 0.00070788 | 0.00070788 |
| 3 | 311.jpg | 0 | 2 | 0 | 1532 | 4763 | 3064 | 1190 | 766 | 2058 | 0.09712128 | 0.00070788 | 0.09782916 |
| 4 | 311.jpg | 0 | 3 | 0 | 2298 | 4763 | 3064 | 1190 | 766 | 655 | 0.030910807 | 0.09782916 | 0.12873997 |
| 5 | 311.jpg | 1 | 0 | 1190 | 0 | 4763 | 3064 | 1190 | 766 | 0 | 0 | 0.12873997 | 0.12873997 |
| 6 | 311.jpg | 1 | 1 | 1190 | 766 | 4763 | 3064 | 1190 | 766 | 0 | 0 | 0.12873997 | 0.12873997 |
| 7 | 311.jpg | 1 | 2 | 1190 | 1532 | 4763 | 3064 | 1190 | 766 | 2632 | 0.12420953 | 0.12873997 | 0.2529495 |
| 8 | 311.jpg | 1 | 3 | 1190 | 2298 | 4763 | 3064 | 1190 | 766 | 2220 | 0.1047664 | 0.2529495 | 0.3577159 |
| 9 | 311.jpg | 2 | 0 | 2380 | 0 | 4763 | 3064 | 1190 | 766 | 0 | 0 | 0.3577159 | 0.3577159 |
| 10 | 311.jpg | 2 | 1 | 2380 | 766 | 4763 | 3064 | 1190 | 766 | 1117 | 0.052713543 | 0.3577159 | 0.41042945 |
| 11 | 311.jpg | 2 | 2 | 2380 | 1532 | 4763 | 3064 | 1190 | 766 | 3048 | 0.14384143 | 0.41042945 | 0.55427086 |
| 12 | 311.jpg | 2 | 3 | 2380 | 2298 | 4763 | 3064 | 1190 | 766 | 3187 | 0.15040113 | 0.55427086 | 0.704672 |
| 13 | 311.jpg | 3 | 0 | 3570 | 0 | 4763 | 3064 | 1190 | 766 | 0 | 0 | 0.704672 | 0.704672 |
| 14 | 311.jpg | 3 | 1 | 3570 | 766 | 4763 | 3064 | 1190 | 766 | 4106 | 0.19377065 | 0.704672 | 0.8984426 |
| 15 | 311.jpg | 3 | 2 | 3570 | 1532 | 4763 | 3064 | 1190 | 766 | 1040 | 0.049079753 | 0.8984426 | 0.9475224 |
| 16 | 311.jpg | 3 | 3 | 3570 | 2298 | 4763 | 3064 | 1190 | 766 | 1112 | 0.052477583 | 0.9475224 | 1 |

Tab[2] Table of 16 quadrants corresponding to the image in Fig[7] with probabilities ("prob"), sum of weights ("sum_w") and continuous interval from 0 to 1 ("prob_from", "prob_to") into which randomly sampled numbers determine which quadrant will be used for sampling.

The following results Fig[8] and Fig[9] were achieved during a test phase with test-time augmentation (TTA) included. For a better alignment of the last layers (responsible for prediction) on inference imagery, we fine-tuned object detector on a manually annotated subset of inference images. The reason for this is that inference and training image sets were not of the same type. Particularly, they came from different sources (Google Earth - CNES/Airbus and xView - DigitalGlobe), consequently their pixel value normalization was different and they did not cover the same geographic locations. To speed up the manual annotation process, we implemented an auto-annotation technique and repeated two loops of this process as described below:

1. We took a small subset of inference images without annotations and ran inference step on them with currently best object detection model
2. We inspected any detected annotations, manually corrected the incorrect annotations and added missing annotations
3. We fine-tuned the best detection model with newly annotated set in order to arrive at a new version of detection model
4. We repeated the entire loop with another subset of inference images without annotations

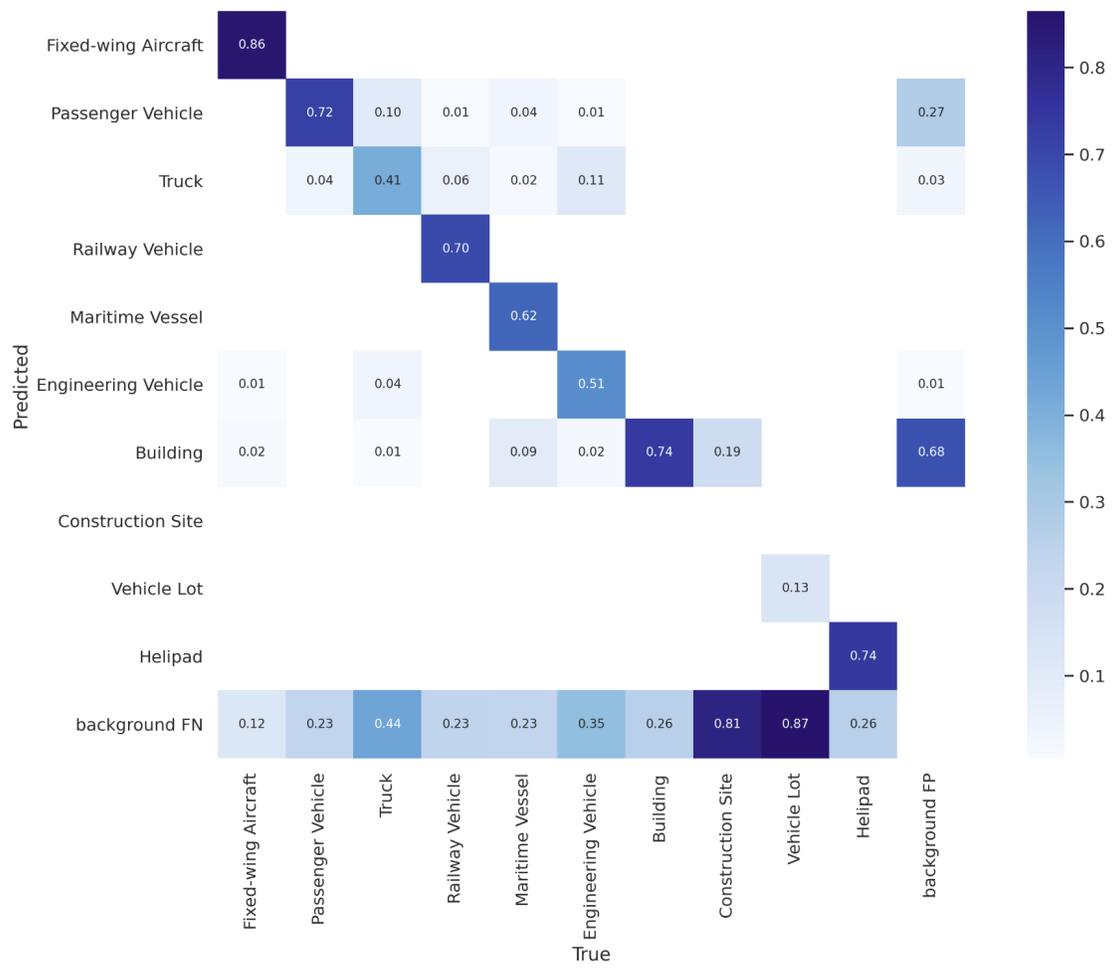

Fig[8] Confusion matrix from Yolov5 framework[13].

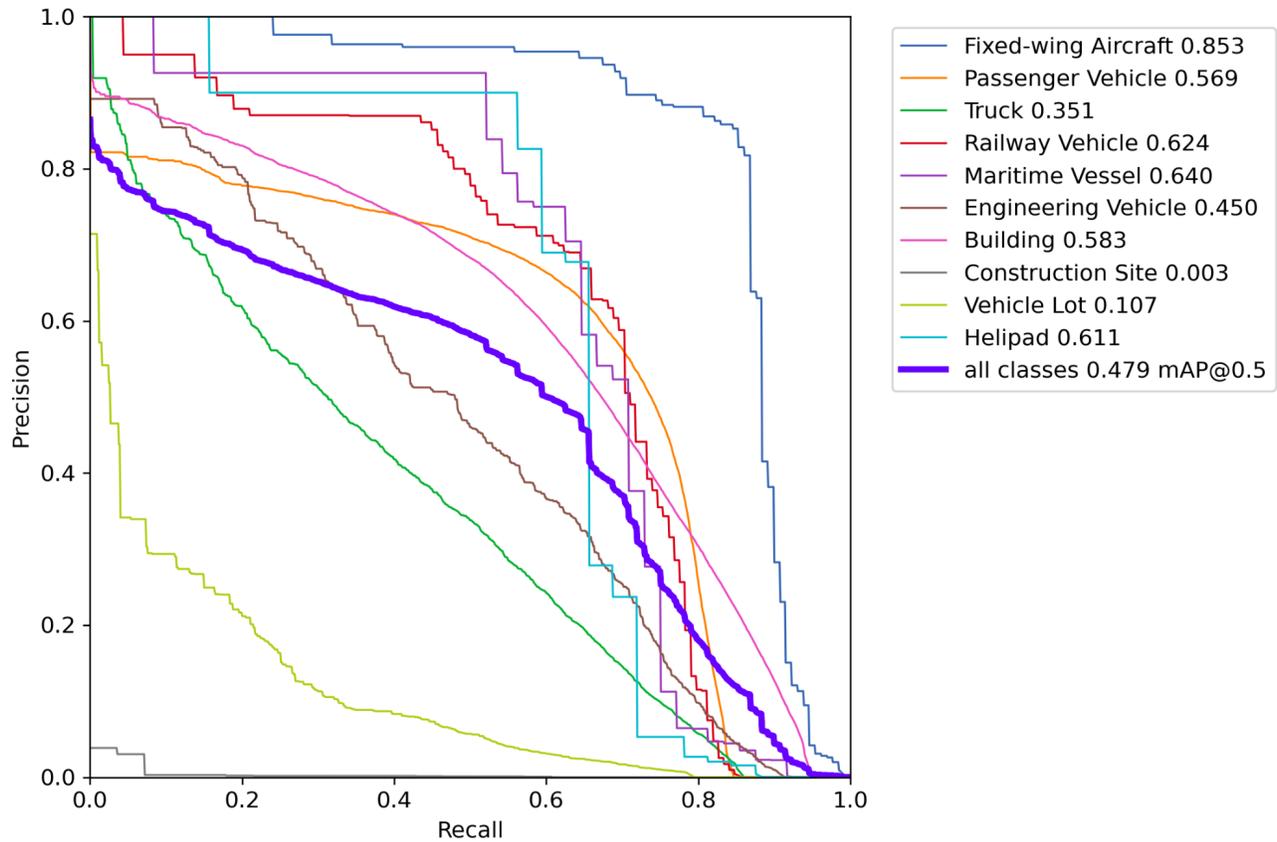

Fig[9] PR-Curve from Yolov5 framework[13]. Performance of each object class in Average Precision (AP) during detection on holdout test set.

# Experiments

The following experiments were conducted:

1. Experiments with low resolution imagery and various nightlight data sets. Tab[3]

| id | description | training_chips | nightlights | architecture | loss_function | valid_loss | error_rate | poverty_valid_r2_ridge |
|---|---|---|---|---|---|---|---|---|
| cls212 | Sentinel2 full_country 10m/px res=384 batchsize=16 | 13000 | VNL annual composite 2015 v1 | Resnet 34 | CrossEntropy Loss | 0.200161 | 0.059484 | 0.4229 |
| cls319 | Sentinel2 full_country 10m/px res=384 batchsize=16 | 13000 | VNL annual composite 2015 v2 | Resnet 34 | CrossEntropy Loss | 0.378415 | 0.1198 | 0.4467 |

Tab[3]

2. Experiments with high resolution imagery and various model sizes, Tab[4].

| id | description | training_chips | nightlights | architecture | loss_function | valid_loss | error_rate | fbeta | poverty_valid_r2_ridge | poverty_valid_rmse_ridge |
|---|---|---|---|---|---|---|---|---|---|---|
| cls103 | GoogleEarth AOI_polygons 0.9m/px res=512 batchsize=8 | 30000 | VNL annual composite 2015 v1 | Resnet 34 | CrossEntropy Loss | 0.483244 | 0.220647 | 0.781344 | 0.7198 | 0.068 |
| cls003 | GoogleEarth AOI_polygons 0.9m/px res=512 batchsize=8 | 30000 | VNL annual composite 2015 v2 | Resnet 152 | CrossEntropy Loss | 0.477439 | 0.203674 | 0.793096 | 0.6502 | 0.076 |

Tab[4]

3. Experiments with regression model, Tab[5]

| id | description | training_chips | nightlights | architecture | loss_function | valid_loss | rmse | poverty_valid_r2_ridge | poverty_valid_rmse_ridge |
|---|---|---|---|---|---|---|---|---|---|
| reg425 | GoogleEarth AOI_polygons 0.9m/px res=512 batchsize=8 | 30000 | VNL annual composite 2015 v2 | Resnet 34 | L1LossFlat | 1.574986 | 3.975069 | 0.492 | 0.0916 |

Tab[5]

4. Experiments with different object detection model sizes and confidence thresholds, Tab[6]

| id | description | training_chips | architecture | loss_function | tta | mAP.5 | poverty_valid_r2 |
|---|---|---|---|---|---|---|---|
| det195 | PyTorch GoogleEarth AOI_polygons 0.3m/px res416 batchsize=16 conf=0.4 | 30000 | Yolov5l | cls_giou_obj | yes | 0.479 | 0.5547 |
| det195 | PyTorch GoogleEarth AOI_polygons 0.3m/px res416 batchsize=16 conf=0.5 | 30000 | Yolov5l | cls_giou_obj | yes | 0.479 | 0.5617 |

Tab[6] Presented r-squared poverty prediction results of object detectors was reached with different confidence thresholds 0.4 and 0.5 and with relative counts of object classes (absolute numbers for each of 10 object classes were divided by the absolute count of trucks)

Four experimental groups should not be compared together, because different spatial resolution of daytime imagery impacts different nightlight values and therefore the subsequent binning into nightlight classes had to be done separately for each experiment group. Different loss functions were defined for them too.

In addition, we performed experiments with modeling techniques for classification models including mixup, label smoothing, progressive resizing and we also trained object detectors with architectures Yolov5m and Yolov5x however none of those brought any significant improvement.

The three models highlighted in yellow proved to have the highest prediction performance on the holdout test set when applied to one final poverty model (considering highest r_squared value out of all other "ensemble models" combinations), Tab[7].

| ensemble_models | ridge_regression_valid_r2 | xgboost_valid_r2 | random_forest_valid_r2 |
|---|---|---|---|
| cls003 / reg425 / det195_conf05 | 0.5007 | 0.768 | 0.6443 |
| cls103 / cls319 / det195_conf05 | 0.8964 | 0.9099 | 0.7411 |
| cls103 / cls212 / det195_conf05 | 0.9096 | 0.9029 | 0.7756 |
| cls103 / cls319 / det195_conf04 | 0.908 | 0.9082 | 0.7151 |
| cls103 / cls212 / det195_conf04 | 0.9037 | 0.898 | 0.7806 |

Tab[7] "Ensemble model" results

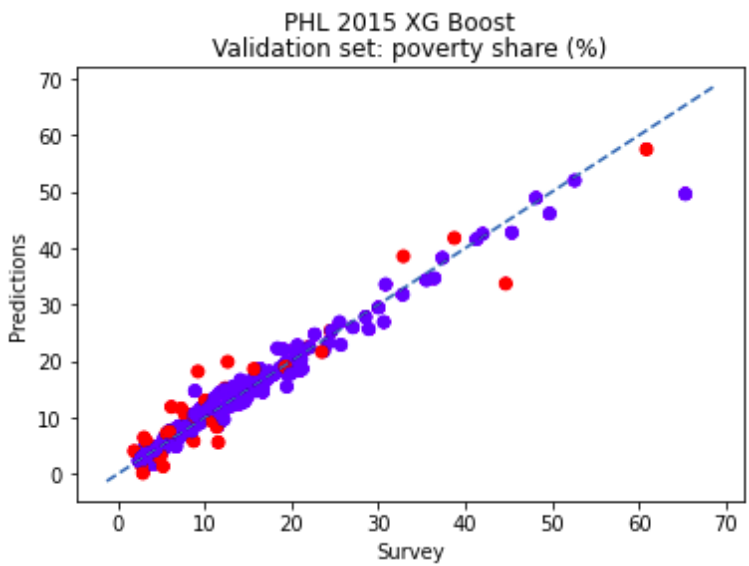

Tab[8] Our best "ensemble model", poverty rates in provinces from validation set in red, from training set in blue

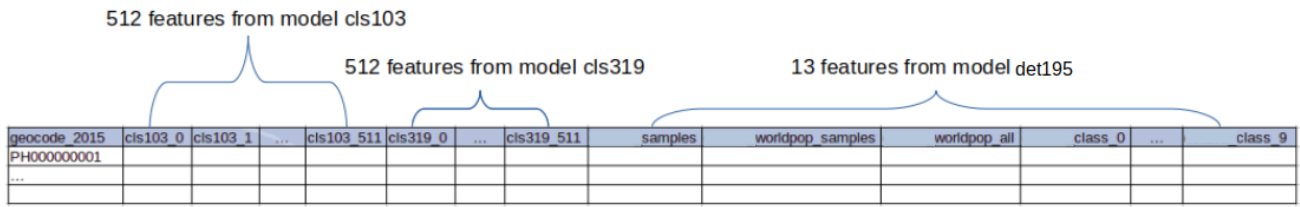

Fig[10] The "ensemble" represents a model trained on the data set created by simple concatenation of extracted feature vectors from each individual model, corresponding to specific geocode of the province. Features for the object detector contained additional 3 features - number of image samples used for specific province, corresponding population within the area of samples for specific province and corresponding population within the whole province based on Worldpop data.

## Results

Results for each experiment group can be seen in Tab[4], Tab[5] and Tab[6]. The models with the highest r-squared values are highlighted in yellow. During linear poverty modeling, we applied ridge regression, xgboost and random forest methods and we took the model with the highest r-squared value.

An "ensemble data set" was created from three best performing models by simple concatenation of columns of extracted features and counts of objects corresponding to the same province (see Fig[10]). The final "ensemble model" was trained on this wide and short data set with method xgboost (gaining the highest r-squared value out of all three methods). To ensure greater confidence in the results, a five fold cross-validation was performed as well.

We think that the best ensemble model has reached high prediction performance r-squared = 0.8289 at province level (calculated as average r_squared value after 5 fold cross-validation) due to the combination of different models and approaches, in this case:

1. natural and man-made features of the studied area at low resolution – Sentinel imagery with 10m/pixel and 3840x3840m GSD

2. natural and man-made features of the studied area at very high resolution – Google Earth imagery with 0.9m/pixel and 450x450m GSD

3. counts of individual object instances for 10 object classes detected at very high resolution – Google Earth imagery with 0.3m/pixel and 450x450m GSD

# Conclusion

Following a literature review, the team believes that no other work employing a similar approach has yet been published. In this paper we attempted to optimize existing poverty prediction approaches, we also experimented with various training methods and presented our ensemble model.
Mostly freely available data, software and hardware were presented in this work with the only exception of Philippines poverty data for a year 2015, which is proprietary data set of ADB. This fact enables national statistical offices to conduct similar modeling at low costs, however still preserving decent model accuracy.

For future work, research spanning over the whole geographic area of the country should be conducted.

# Acknowledgments

Poverty data used in this study are based on the information provided on Mapping Poverty through Data Integration and Artificial Intelligence: A Special Supplement of the Key Indicators for Asia and the Pacific published by Asian Development Bank (2020).

Essential were also all publicly available data sets presented, including CNES/Airbus, xView, VIIRS, Sentinel and Worldpop data sets. Big acknowledgement goes to Google company for making Google Earth data, Earth Engine and Google Colab publicly available. We would also like to acknowledge Fastai and Ultralytics teams for their great work in the field of open software.

We would like to acknowledge all the reviewers who gave us insightful comments of this work, especially Mr. Martin Hofer and Mr. Kristofer Hamel.

# Abbreviations

NGO - non-governmental organisation
VIIRS - The Visible Infrared Imaging Radiometer Suite
GSD - ground sampling distance
VNL - VIIRS nighttime light
GPU - Graphics processing unit
ADB - Asian Development Bank
CNES - Centre National d'Etudes Spatiales (CNES), France
VHR - very high resolution satellite imagery, 0.3m per pixel in this article
TIF - Tagged Image Format used in geolocated imagery
ETL - extract-transform-load